\documentclass[letterpaper, 10 pt, conference]{ieeeconf}

\IEEEoverridecommandlockouts 
\usepackage{amsmath}
\usepackage{amssymb}
\usepackage{color,soul}
\usepackage{multirow}
\usepackage[pdftex]{graphicx}\usepackage[colorlinks]{hyperref}
\usepackage{url}
\usepackage[linesnumbered,ruled,vlined]{algorithm2e}
\SetArgSty{textnormal}
\usepackage[belowskip=-10pt,aboveskip=5pt]{caption}
\usepackage{caption}
\usepackage{subcaption}

\begin{document}

\title{The Role of Tactile Sensing in Learning and Deploying \\Grasp Refinement Algorithms}

\author{{Alexander Koenig$^{1,2}$, Zixi Liu$^{2}$, Lucas Janson$^{3}$ and Robert Howe$^{2,4}$}
\thanks{This material is based upon work supported by the US National Science Foundation under Grant No. IIS-1924984 and by the German Academic Exchange Service.}
\thanks{$^{1}$ Department of Informatics, Technical University of Munich}
\thanks{$^{2}$ School of Engineering and Applied Sciences, Harvard University}
\thanks{$^{3}$ Department of Statistics, Harvard University}
\thanks{$^{4}$ RightHand Robotics, Inc., 237 Washington St, Somerville, MA 02143 USA. Robert Howe is corresponding author \texttt{howe@seas.harvard.edu}.}
}
\maketitle

\begin{abstract}
A long-standing question in robot hand design is how accurate tactile sensing must be. This paper uses simulated tactile signals and the reinforcement learning (RL) framework to study the sensing needs in grasping systems.
Our first experiment investigates the need for rich tactile sensing in the rewards of RL-based grasp refinement algorithms for multi-fingered robotic hands. We systematically integrate different levels of tactile data into the rewards using analytic grasp stability metrics. We find that combining information on contact positions, normals, and forces in the reward yields the highest average success rates of 95.4\% for cuboids, 93.1\% for cylinders, and 62.3\% for spheres across wrist position errors between 0 and 7 centimeters and rotational errors between 0 and 14 degrees. This contact-based reward outperforms a non-tactile binary-reward baseline by 42.9\%.
Our follow-up experiment shows that when training with tactile-enabled rewards, the use of tactile information in the control policy's state vector is drastically reducible at only a slight performance decrease of at most 6.6\% for no tactile sensing in the state. Since policies do not require access to the reward signal at test time, our work implies that models trained on tactile-enabled hands are deployable to robotic hands with a smaller sensor suite, potentially reducing cost dramatically.
\end{abstract}

\IEEEpeerreviewmaketitle

\section{Introduction}

Tactile sensing provides essential information about local object geometry, surface properties, contact forces, and grasp stability \cite{Cutkosky2016}. Hence, tactile sensors can be a valuable tool in robotic grasp refinement tasks \cite{dollar2010contact} where a grasping system recovers from calibration errors. Computer vision approaches for grasp refinement often face limitations due to the occlusion of contact events. Tactile sensors are often expensive and fragile hardware components. Hence, for cost-effective robotic hand design, it is essential to understand when robot hands need precise sensing and how accurate it should be to achieve good grasping performance. 

A few research papers investigated the effect of tactile sensor resolution on grasp success. Wan et al. \cite{Wan2018} found that reduced spatial resolution of tactile sensors negatively impacts grasp success since inaccuracies in contact position and normal sensing can influence grasp stability predictions. Other works analyzed the effect of contact sensor resolution on grasp performance in the context of reinforcement learning. In simulated experiments, Merzić et al. \cite{merzic2019contact} found that contact feedback in a policy's state vector improves the performance of RL-based grasping controllers, and \cite{melnik2019tactile, melnik2021} presented similar results for in-hand manipulation. However, \cite{melnik2019tactile, melnik2021} also concluded that models trained with binary contact signals perform equally well as models that receive accurate normal force information. Furthermore, \cite{melnik2019tactile, melnik2021} found that tactile resolution (92 vs. 16 sensors) has no noticeable effect on performance and sample efficiency of reinforcement learned manipulation controllers.

\begin{figure}
	\includegraphics[width=\linewidth]{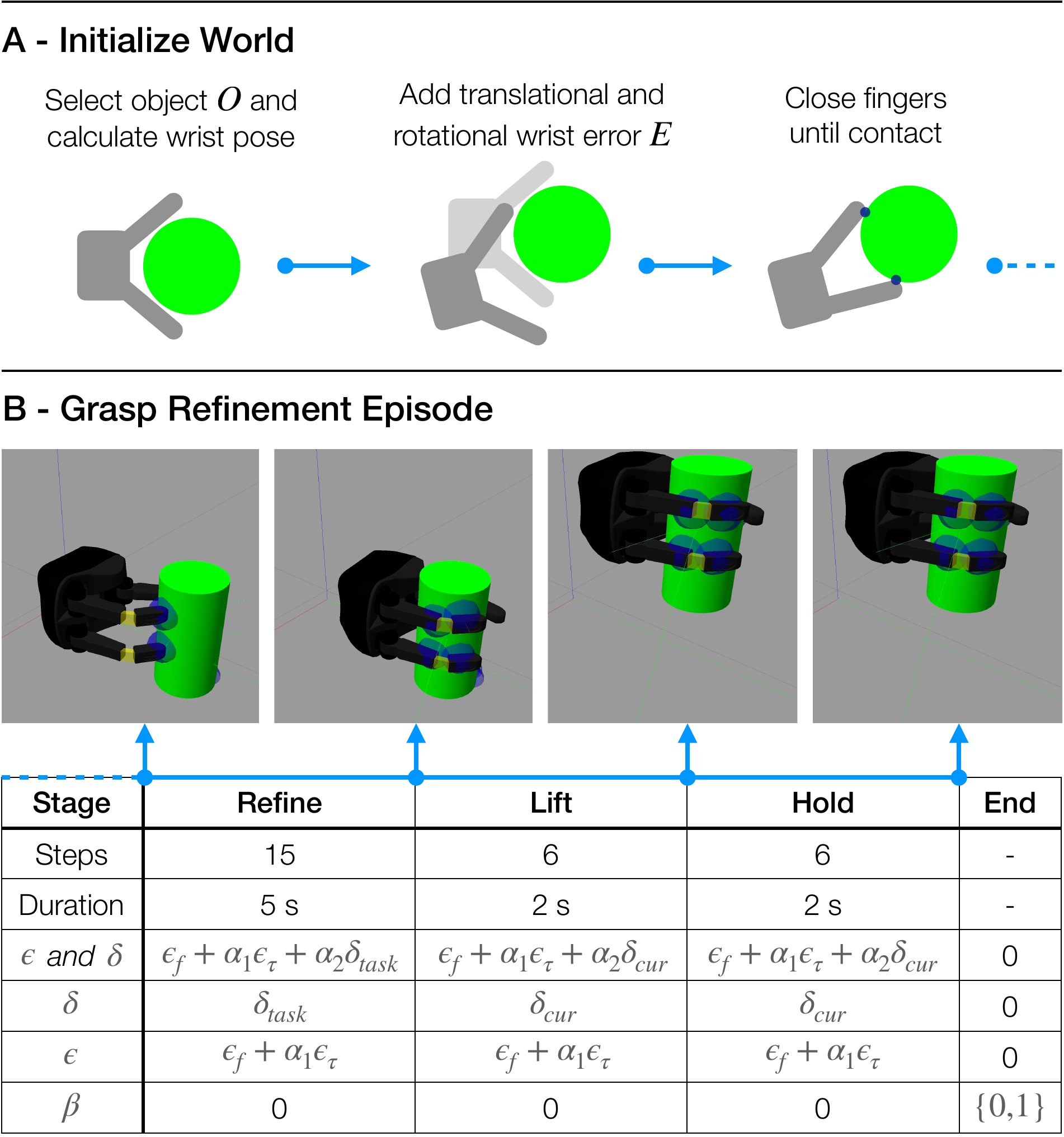}
	\caption{Overview of one algorithm episode. (A) Initialization of hand and object. (B) We split the grasp refinement algorithm into four stages and compare four reward frameworks: (1) $\epsilon$ \textit{and} $\delta$, (2) only $\delta$, (3) only $\epsilon$ and (4) the non-tactile binary reward baseline $\beta$. The weighting factors of $\alpha_1=5$ and $\alpha_2=0.5$ were empirically determined.}
	\label{fig:overview}
	\vspace{-0.1cm}
\end{figure}

\begin{table}[h]
\captionsetup{skip=5pt, position = bottom} 
\caption{Reward functions of RL grasping controllers.}
\vspace{0.5cm} 
\begin{tabular}{p{0.12\linewidth}|p{0.77\linewidth}}
\textbf{Paper}  & \textbf{Reward}  \\ \hline
Chebotar \newline 2016 \cite{chebotar2016spatio-temporal} &
\underline{Maximize} predicted grasp success from learned stability predictor \\
\hline 
Merzić \newline 2019 \cite{merzic2019contact} &
\underline{Maximize} (1) number of links in contact and (2) binary drop test reward \newline \underline{Minimize} (1) distance object to gripper, (2) distance fingertips to object, (3) joint torques and (4) object velocity  \\   
\hline 
Wu \newline 2019 \cite{wu2019mat} & 
\underline{Maximize} binary pick-up reward at episode end \newline \underline{Minimize} finger reopening \\     
\hline 
Hu \newline 2020 \cite{hu2020reaching} & 
\underline{Maximize} (1) number of contact points and (2) number of object key-points contained in convex hull of hand and finger key-points \newline
\underline{Minimize} (1) distance from hand key-points to object key-points, (2) angle between hand key-point normals and vectors pointing from hand key-points to object center, (3) number of contacts on outside of fingers and (4) object linear velocity 
\\           
\end{tabular}
\vspace{-0.5cm} 
\label{tab:related_works}
\end{table}

In this paper, we use accurate tactile signals from simulation and the reinforcement learning framework to explore the tactile sensing needs in robotic systems. RL algorithms aim to produce a policy $\pi(\boldsymbol{a}|\boldsymbol{s})$ that outputs actions $\boldsymbol{a}$ given state information $\boldsymbol{s}$ such that the cumulative reward signal $r$ is maximized. The reward function is a critical part of every RL algorithm \cite{silver2021reward}. While the previous work in \cite{merzic2019contact, melnik2019tactile, melnik2021} only studied the tactile resolution in the policy's state, our first contribution investigates the impact of tactile information in the reward signal. 
Table \ref{tab:related_works} shows an overview of the reward functions used in other RL-based grasping controllers that process tactile information from multi-fingered hands. These reward functions are insufficient to study the effect of different types of contact information, because they either directly encode the experiment outcome \cite{merzic2019contact, wu2019mat} or consist of manually engineered cues (e.g., number of contacts \cite{merzic2019contact, hu2020reaching}) that do not include contact position, normal, and force information. Hence, we propose a unified framework to systematically incorporate different levels of tactile information from robotic hands into a reward signal via analytic grasp stability metrics. As shown in Fig. \ref{fig:overview}, we conduct grasp refinement experiments and define three types of rewards: $\epsilon$ calculated from contact positions and normals, a force-based reward $\delta$, and a binary task execution reward $\beta$. By comparing the individual and combined performance of $\epsilon$ and $\delta$, we estimate the relevance of contact position, normal, and force sensing for the reward signal.

\begin{figure}
\vspace{0.2cm} %
\centering
	\includegraphics[width=\linewidth]{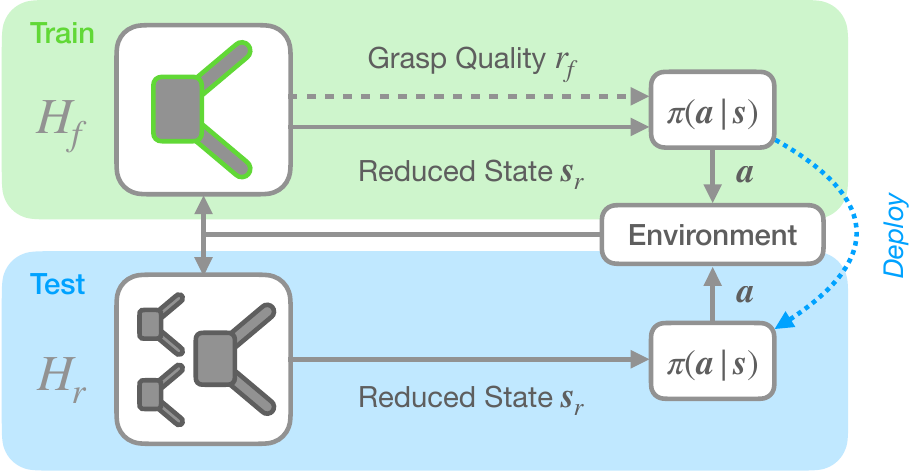}
	\caption{The hypothesized workflow for training and deploying RL-controlled grasping systems. First, train a policy $\pi(\boldsymbol{a}|\boldsymbol{s})$ on a hand $H_f$ with a \textit{full} tactile sensor suite (e.g., contact position, normal and force sensors) where the grasp quality metrics are available as a reward $r_f$ to learn a task, but only provide a subset of the available contact data in the state vector $\boldsymbol{s}_r$. Afterwards, deploy the policy to many structurally similar hands $H_r$ with a \textit{reduced} sensor set to save cost.}
	\label{fig:c2_flow}
\end{figure}

Calculating grasp stability metrics requires costly tactile sensing capabilities on physical grippers. However, the reward signal is only required during the training of policies but not while testing, which suggests that sensing needs in both stages could be different. We hypothesize in Figure \ref{fig:c2_flow} that policies trained with grasp stability metrics on a robotic hand $H_f$ with a \textit{full} tactile sensor suite are deployable to structurally similar but more affordable hands $H_r$ with \textit{reduced} tactile sensing at a small performance decrease. Hence, our second experiment gradually decreases tactile resolution in the state vector to find realistic training and deployment workflows for grasping algorithms.

Our paper reviews and defines grasp quality metrics in section \ref{sec:metrics}. We use these metrics as reward signals in section \ref{sec:reward_design} and thereby study the impact of contact position, normal, and force sensing in the reward. In section \ref{sec:contact_sensing}, we investigate the importance of tactile data in the state vector.

\section{Grasp Stability Metrics\label{sec:metrics}}

\subsection{Largest-minimum resisted forces and torques}


Ferrari and Canny \cite{ferrari1992epsilon} define grasp quality as the largest-minimum perturbing wrench that the grasp can resist given the grasp's force constraints. Ferrari's metric \cite{ferrari1992epsilon} suffers from the non-comparability of forces (in N) and torques (in Nm). Hence, Mirtich and Canny \cite{mirtich1994easily} refine this popular metric by decoupling the wrench space into a force and torque component, and thereby evaluate how well a grasp resists pure forces and torques. 

\begin{figure}[h]
	\includegraphics[width=\linewidth]{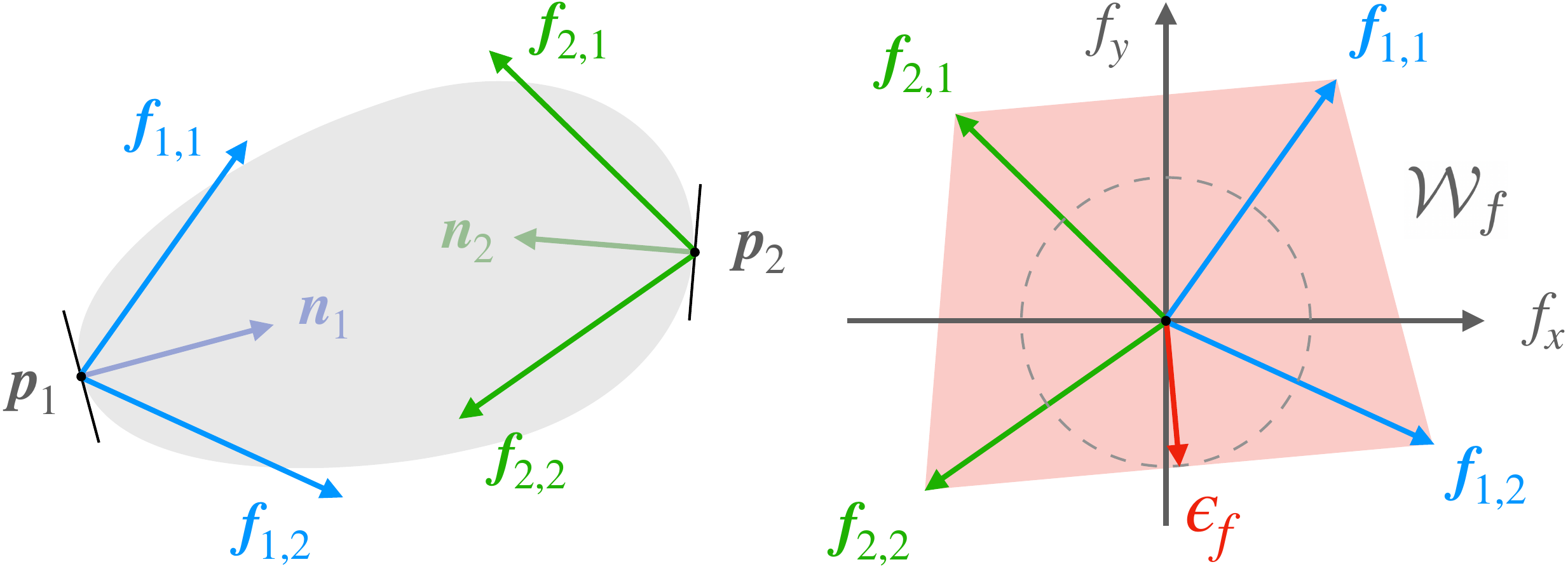}
	\caption{Left: a grasp on a grey object with two contact points $\boldsymbol{p}_1$ and $\boldsymbol{p}_2$, contact normals $\boldsymbol{n}_i$ and friction cones. Right: the quality metric $\epsilon_f$ is the radius of the largest ball contained in the convex hull $\mathcal{W}_f$ over the set of resisted forces.}
	\label{fig:epsilon_f}
\end{figure}

Let us examine how to measure resistance to disturbing forces. The contact force $\boldsymbol{f}_i$ at each contact $i$ is constrained via the friction cone ${f}_{i,t} \leq \mu {f}_{i,n}$, where $\mu$ is the coefficient of friction and ${f}_{i,t}$ and ${f}_{i,n}$ are the tangential and normal components of $\boldsymbol{f}_i$, respectively. The friction cone is discretized using $m$ edges $\boldsymbol{f}_{i,j}$. The set of forces $\mathcal{W}_{f}$ that the contacts can apply to the object is $\mathcal{W}_{f}=\operatorname{ConvexHull}\left(\bigcup_{i=1}^{n}\left\{\boldsymbol{f}_{i, 1}, \dots, \boldsymbol{f}_{i, m}\right\}\right)$, where $n$ is the number of contacts. Finally, the quality metric $\epsilon_f$ in equation \eqref{eq:epsilon} is the shortest distance from the origin to the nearest hyper-plane of $\mathcal{W}_f$. Hence, the metric defines a lower bound on the resisted force in all directions. As shown in Fig. \ref{fig:epsilon_f}, $\epsilon_f$ can be geometrically interpreted as the radius of the largest ball centered at the origin and contained inside $\mathcal{W}_f$.

\begin{equation}
\label{eq:epsilon}
	\epsilon_f =\min_{\boldsymbol{f} \in \partial \mathcal{W}_f}\|\boldsymbol{f}\|
\end{equation}

This concept is easily extended to the torque domain. The reaction torque $\boldsymbol{\tau}_{i,j}$ resulting from a friction cone edge $\boldsymbol{f}_{i, j}$ is calculated by $\boldsymbol{\tau}_{i, j}=\boldsymbol{r_{i}} \times \boldsymbol{f_{i, j}}$, where $\boldsymbol{r_{i}}$ is a vector pointing from the object's center of mass to the contact point $\boldsymbol{p_i}$. The set of torques $\mathcal{W}_{\tau}$ that the grasp can resist is defined by $\mathcal{W}_{\tau}=\operatorname{ConvexHull}\left(\bigcup_{i=1}^{n}\left\{\boldsymbol{\tau}_{i, 1}, \dots, \boldsymbol{\tau}_{i, m}\right\}\right)$. The metric $\epsilon_{\tau}$ in equation \eqref{eq:epsilon_torque} evaluates the grasp's quality by identifying the magnitude of the largest-minimum resisted torque. 

\begin{equation}
\label{eq:epsilon_torque}
	\epsilon_{\tau} =\min_{\boldsymbol{\tau} \in \partial \mathcal{W}_{\tau}}\|\boldsymbol{\tau}\|
\end{equation}

\subsection{Minimum distance to the friction cone}

The quality metrics $\epsilon_f$ and $\epsilon_{\tau}$ analyze the forces that each contact can theoretically exert on the object. However, these metrics do not consider the actual contact forces that the contacts apply to the object. To this end, we define two force-based quality metrics $\delta_{cur}$ and $\delta_{task}$. While $\delta_{cur}$ is a general-purpose grasp quality metric, $\delta_{task}$ is applicable when a task definition exists. 

\begin{figure}[h]
\centering
	\includegraphics[width=0.7\linewidth]{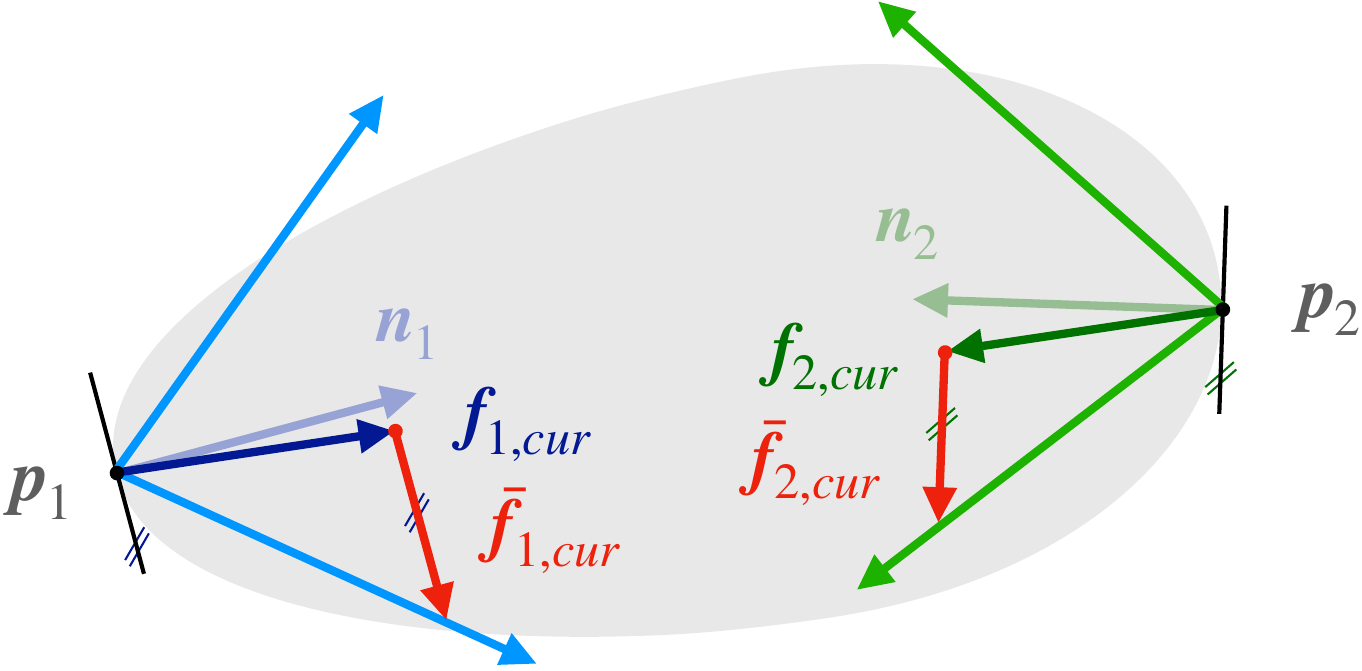}
	\caption{Grasp with current contact forces $\boldsymbol{f}_{i,cur}$ and tangential force margins $\boldsymbol{\bar{f}}_{i,cur}$ to the friction cones.}
	\label{fig:delta_cur}
\end{figure}

Similar to Buss et al. \cite{buss1996forceoptim}, we measure grasp stability in terms of how far the contact forces are from the friction limits. Fig. \ref{fig:delta_cur} shows a grasp with the current contact forces $\boldsymbol{f}_{i,cur}$ and the tangential force margins $\boldsymbol{\bar{f}}_{i,cur}$. The vectors $\boldsymbol{\bar{f}}_{i,cur}$ are forces in the tangential direction that point from $\boldsymbol{f}_{i,cur}$ to the closest point on the friction cone, thereby identifying the direction in which the contact can take the least tangential force before slipping. A grasp with large tangential force margins $\boldsymbol{\bar{f}}_{i,cur}$ is desirable since the contacts are less prone to sliding when an object wrench is applied. Hence, the metric $\delta_{cur}$ in equation \eqref{eq:delta_cur} measures the average magnitude of the safety margins $\| \boldsymbol{\bar{f}}_{i,cur} \|$. Contacts with larger forces contribute more to grasp stability because they can have larger tangential force margins  $\boldsymbol{\bar{f}}_{i,cur}$ and can thereby compensate for more disturbing object wrenches. Therefore, we weigh the safety margins $\| \boldsymbol{\bar{f}}_{i,cur} \|$ by their respective contact force magnitudes $\| \boldsymbol{f}_{i,cur}\|$. 

\begin{equation}
\label{eq:delta_cur}
\delta_{cur} = \frac{\sum_{i=1}^{n_c} \|\boldsymbol{f}_{i,cur} \| \|\boldsymbol{\bar{f}}_{i,cur}\|}{\sum_{i=1}^{n_c} \|\boldsymbol{f}_{i,cur}\|}
\end{equation}

In many grasping tasks, a clear task definition exists. Let $T = \{\boldsymbol{w_1}, \boldsymbol{w_2}, \dots, \boldsymbol{w_m}\}$ be the set of task wrenches that the grasp must resist during task execution (e.g., object weight or wrenches from expected collisions). Several task-oriented quality metrics measure how well a convex set of $T$ is contained within the convex set of all wrenches that the grasp can resist \cite{li1988task, pollard1996, borst2004}. However, since these approaches reason about the theoretically applicable contact forces, which are commonly bounded to unity \cite{roa2015grasp, ferrari1992epsilon}, it is not possible to evaluate whether the \textit{current} contact forces of a grasp are suitable to balance the anticipated task wrenches. 

\begin{figure}[h]
\centering
	\includegraphics[width=0.7\linewidth]{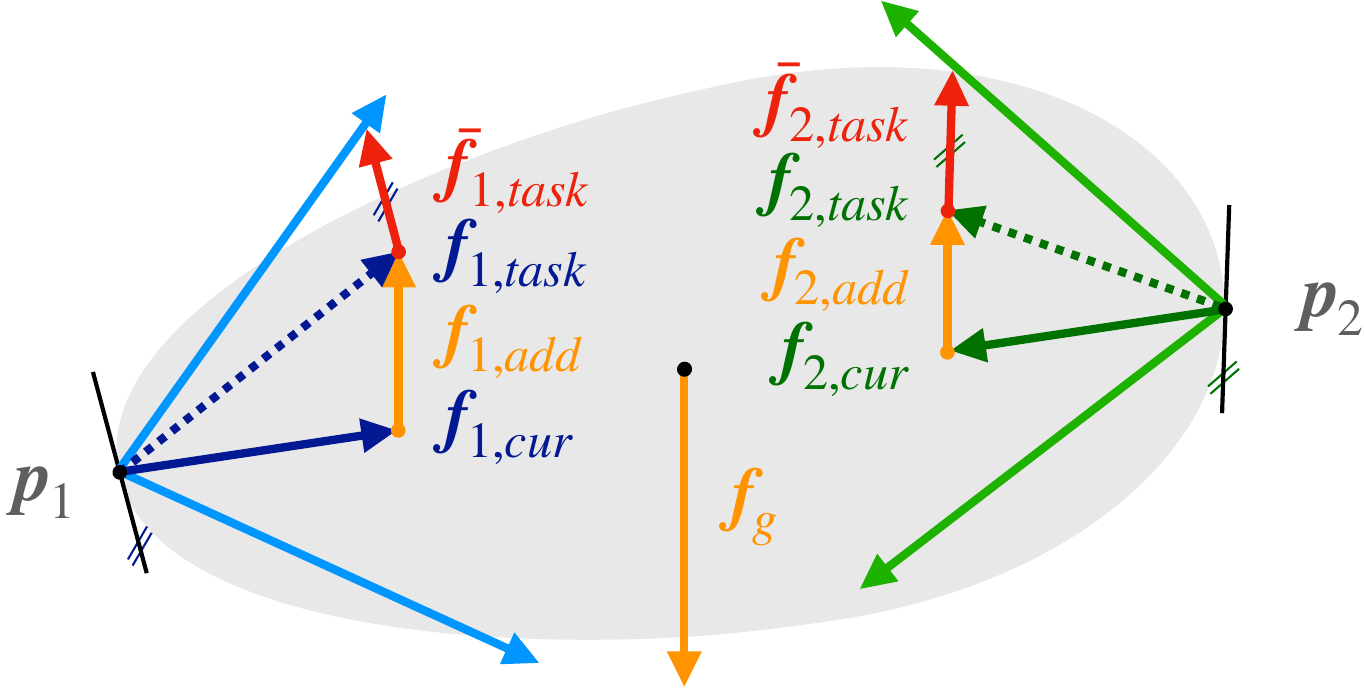}
	\caption{Grasp with predicted task contact forces $\boldsymbol{f}_{i,task}$ after mapping the task force $- \boldsymbol{f}_{g}$ onto the contacts.}
	\label{fig:delta_task}
\end{figure}

We define an alternative task-oriented metric $\delta_{task}$. We calculate the additional contact force $\boldsymbol{f}_{i, add}$ that each contact $i$ must react with to compensate the task wrench $\boldsymbol{w} \in T$ with $\mathbf{G}^{+} \boldsymbol{w} = (\begin{array}{cccc}  \boldsymbol{f}_{1, add}^T&\boldsymbol{f}_{2, add}^T&\dots&\boldsymbol{f}_{n_c, add}^T \end{array})^T$, where $\mathbf{G}^{+}$ is  the pseudoinverse of the grasp matrix as defined in \cite{Prattichizzo2008}. Fig. \ref{fig:delta_task} shows that the task contact force $\boldsymbol{f}_{i,task} = \boldsymbol{f}_{i,cur} + \boldsymbol{f}_{i,add}$ is the sum of the current contact force $\boldsymbol{f}_{i, cur}$ and $\boldsymbol{f}_{i, add}$ which results from a task wrench (here the object weight $-\boldsymbol{f}_{g}$). We use a hard contact model and assume that the internal grasp forces stay the same after applying $\boldsymbol{f}_{i, add}$. The metric $\delta_{task}$ in equation \eqref{eq:delta_task} measures the expected grasp stability during task execution by computing the average magnitude of the tangential force margins $\| \boldsymbol{\bar{f}}_{i,task}\|$ of the task contact forces $\boldsymbol{f}_{i,task}$. The metric $\delta_{task}$ is a lower bound over all task wrenches $\boldsymbol{w}\in T$ and we thereby identify the worst-case task wrench.

\begin{equation}
\label{eq:delta_task}
\delta_{task} =\min_{\boldsymbol{w}\in T} \frac{\sum_{i=1}^{n_c} \|\boldsymbol{f}_{i,task} \| \|\boldsymbol{\bar{f}}_{i,task}\|}{\sum_{i=1}^{n_c} \|\boldsymbol{f}_{i,task}\|}
\end{equation}

\section{Tactile Sensing and the Reward Function\label{sec:reward_design}}

\subsection{Simulation Environment}

We simulate the grasp refinement episodes of the three-fingered ReFlex TakkTile hand (RightHand Robotics, Somerville, MA USA) using a custom robotics simulator based on the Gazebo \cite{gazebo} simulation environment, the DART \cite{dart} physics engine, and the ROS \cite{quigley09} communication framework. We model the under-actuated distal flexure as a rigid link with two revolute joints (one between the proximal and one between the distal finger link). Further, we approximate the finger geometries as cuboids to reduce computational load. We activate simulated gravity in our experiments (unlike in \cite{merzic2019contact}), and the object can freely interact with the hand and the world. Our source code is publicly available under \url{github.com/axkoenig/grasp_refinement}. 

\subsection{Train and Test Dataset}

\begin{figure}[h]
\vspace{0.2cm}
	\centering
	\includegraphics[width=0.7\linewidth]{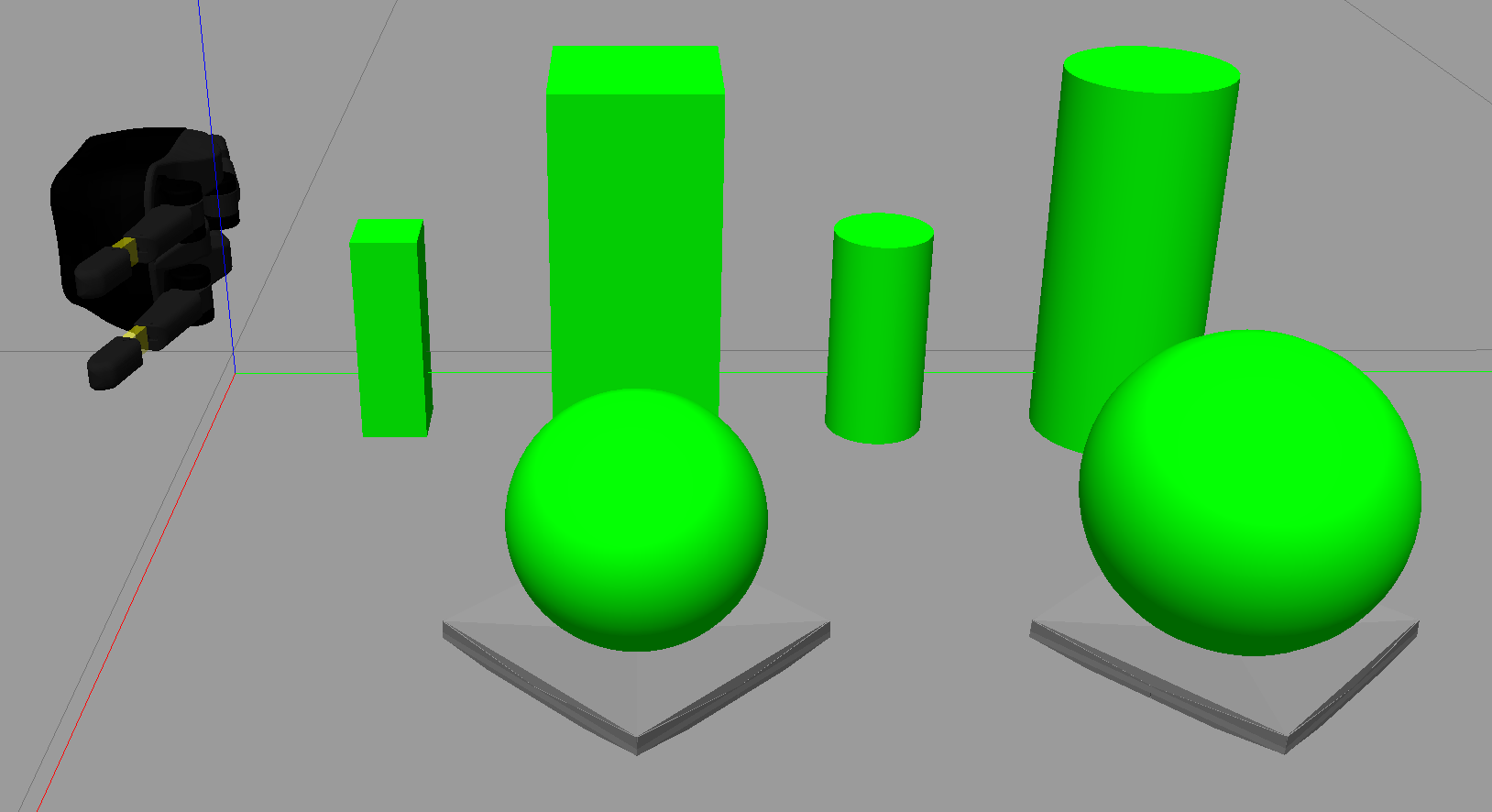}
	\caption{Minimum and maximum object sizes. Sizes in mm: cuboid height $\in [130, 230]$, length and width $ \in [40, 100]$, cylinder height $\in [130, 230]$, radius $ \in [30, 50]$, sphere radius $ \in [65, 80]$. We place the spheres on a concave mount to prevent rolling.}
	\label{fig:objects}
\end{figure}

Each training sample consists of a tuple $(O, E)$, where $O$ is the object, and $E$ is the wrist pose error sampled uniformly before every episode. There are three object types (cuboid, cylinder, and sphere) with a mass $\in [0.1, 0.4]$ kg and randomly sampled sizes. Fig. \ref{fig:objects} visualizes the minimum and maximum object dimensions. 
The wrist pose error $E$ consists of a translational and a rotational error. We uniformly sample the translational error $(e_x,e_y,e_z)$ from $[-5, 5] \text{ cm}$ and the rotational error $(e_\xi, e_\eta, e_\zeta)$ from $[-10, 10]  \text{ deg}$ for each variable, respectively.

\begin{figure}[h]
	\centering
	\includegraphics[width=0.7\linewidth]{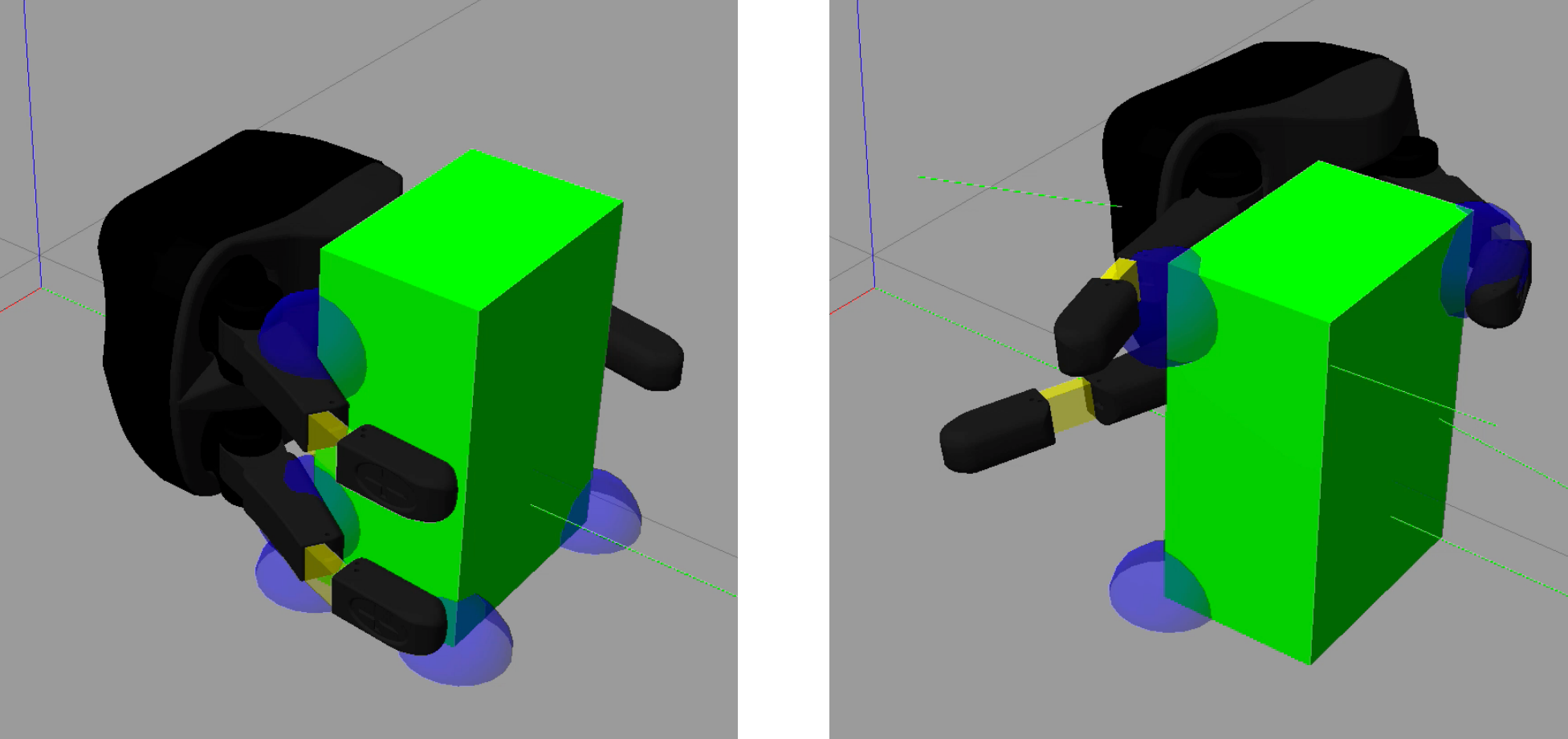}
	\caption{Left: wrist error case \textbf{A} (no wrist error), Right: wrist error case \textbf{H} (maximum wrist error) after closing the fingers. Contact points in blue.}
	\label{fig:wrist_errors}
\end{figure}

We define 8 different wrist error cases for the test dataset. Let $d(a,b,c) = \sqrt{a^2+b^2+c^2}$ be the L2 norm of the variables $(a,b,c)$. Table \ref{table:wrist_errors} shows the wrist error cases, where case A corresponds to no error and case H means maximum wrist error. Fig. \ref{fig:wrist_errors} visualizes two wrist error cases. The test dataset consists of 30 random objects $O$ (10 cuboids, 10 cylinders, and 10 spheres). Per object $O$, we randomly generate the eight wrist error cases $\{A,B, \dots, H \}$ from Table \ref{table:wrist_errors}. Hence, we run $30\times8=240$ experiments to test one model.
\begin{table}[h]
\centering
\caption{Wrist error cases}
\label{table:wrist_errors}
\begin{tabular}{|l|l|l|l|l|l|l|l|l|}
\hline
\textbf{Wrist Error Case}  & \textbf{A} & \textbf{B} & \textbf{C} & \textbf{D} & \textbf{E} & \textbf{F} & \textbf{G} & \textbf{H} \\ \hline
$d(e_x,e_y,e_z)$ in cm           & 0          & 1          & 2          & 3          & 4          & 5          & 6          & 7          \\ \hline
$d(e_\xi, e_\eta, e_\zeta)$ in deg & 0          & 2          & 4          & 6          & 8          & 10         & 12         & 14         \\ \hline
\end{tabular}
\end{table}

\subsection{State and Action Space\label{sec:state_action_space}}

The state vector $\boldsymbol{s}$ consists of 7 joint positions (1 finger separation, 3 proximal bending, 3 distal bending degrees of freedom), and 7 contact cues (3 on proximal links, 3 on distal links, and 1 on palm) that include contact position, contact normal and contact force, which have 3 $(x,y,z)$ components each. The dimension of the state vector is $\boldsymbol{s} \in \mathbb{R}^{7+7\times(3\times3)=70}$. Note that we do not assume any information about the object (e.g., object pose, geometry, or mass) in the state vector. 
The action vector $\boldsymbol{a}$ consists of 3 finger position increments, 3 wrist position increments and 3 wrist rotation increments. The action vector's dimension is $\boldsymbol{a} \in \mathbb{R}^{3+3+3=9}$. 
The policy $\pi_{\boldsymbol{\theta}}$ is parametrized by a neural network with weights $\boldsymbol{\theta}$. The network is a multi-layer perceptron (MLP) with four layers [70, 256, 256, 9]. We use the \texttt{stable-baselines3} \cite{stable-baselines3} implementation of the soft actor-critic (SAC) \cite{haarnoja2018sac} framework to train the stochastic policy $\pi_{\boldsymbol{\theta}}$. We evaluate the policy deterministically when testing.

\subsection{Algorithm Overview}

Fig. \ref{fig:overview} shows an overview of one training episode. Before starting the control algorithm, we reset the world. Thereby, we randomly generate a new object, wrist error tuple $(O,E)$ (or we select one from the test dataset). We assume a computer vision system and a grasp planner that produces a side-ways facing grasp at a fixed  5 cm offset from the object's center of mass. We add the wrist pose error $E$ to this grasp pose to simulate calibration errors and close the fingers of the robotic hand in the erroneous wrist pose until the fingers make contact with the object. Consequently, the grasp refinement episode starts. We divide each episode into three stages, as displayed in Fig. \ref{fig:overview}. Firstly, the policy $\pi_{\boldsymbol{\theta}}$ \textit{refines} the grasp in five seconds and 15 algorithm steps. Afterward, the agent \textit{lifts} the object by 15 cm via hard-coded increments to the wrist's $z$-position in two seconds and six algorithm steps. Finally, the policy \textit{holds} the object in place for two seconds and six algorithm steps to test the grasp's stability. The policy $\pi_{\boldsymbol{\theta}}$ can update the wrist and finger positions while lifting and holding. The control frequency of the policy in all stages is 3 Hz, while the update frequency of the low-level proportional–derivative (PD) controllers in the wrist and the fingers is 100 Hz. 

Each episode can last at most $15+6+6=27$ algorithm steps. We end the episode earlier if the hand shifts the object by more than 10 cm during the refinement stage to discourage excessive movement of the object. Furthermore, we terminate refinement if one of the fingers exceeds a joint limit of 3 radians. We do not enter the holding stage if the object dropped after the lifting stage. The algorithm trains for 25000 steps, which corresponds to approximately 1000 training episodes depending on the episode lengths. 

As shown in the table of Fig. \ref{fig:overview}, we use the analytic grasp stability metrics from section \ref{sec:metrics} as reward functions. We compare the following reward configurations: (1) both $\epsilon$ \textit{and} $\delta$, (2) only $\epsilon$, (3) only $\delta$ and (4) the baseline $\beta$. Fig. \ref{fig:overview} shows that $\delta$ refers to $\delta_{task}$ in the \textit{refine} stage to measure expected grasp stability before lifting and $\delta_{cur}$ in the \textit{lift} and \textit{hold} stages to measure current stability. Further, $\epsilon$ is a weighted combination of $\epsilon_f$ and $\epsilon_{\tau}$. While $\epsilon$ \textit{and} $\delta$, $\delta$, and $\epsilon$ provide stability feedback after every algorithm step, the baseline $\beta$ gives a sparse reward after the holding stage, indicating if the object is still in the hand (1) or not (0). Since the SAC algorithm is sensitive to reward scaling \cite{haarnoja2018sac}, we normalize the rewards, which are based on grasp quality metrics.

\begin{figure}[t]
	\centering
    	\includegraphics[width=\linewidth]{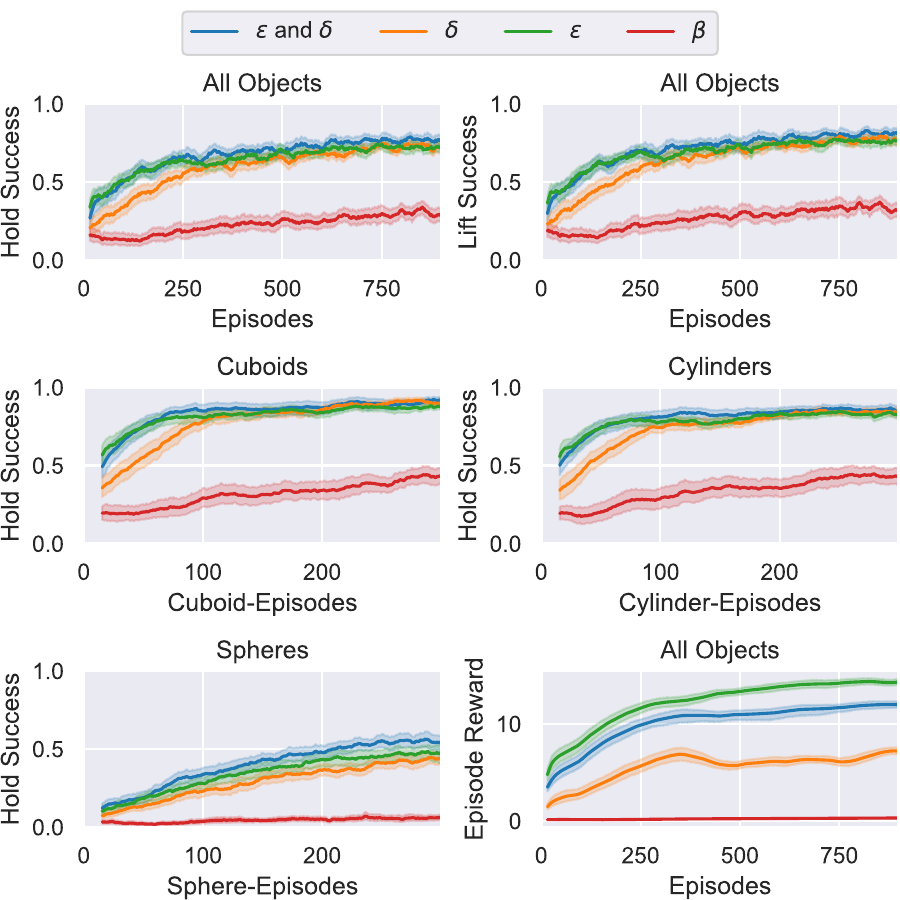}
    	\caption{Training results for reward frameworks.}
    	\label{fig:train_reward_frameworks}
\end{figure}

\subsection{Results}

Fig. \ref{fig:train_reward_frameworks} shows the training results of the four reward frameworks. For all experiments in this paper, we average over 40 models trained with different seeds for each framework and smooth the training curves with a moving average filter of kernel size 30. The error bars in all plots represent $\pm 2$ standard errors. It takes approximately 20 hours to train one model on a machine with 4 CPUs. We realize from Fig. \ref{fig:train_reward_frameworks} that the algorithms trained with grasp stability metrics are more sample efficient and reach higher success rates than $\beta$ within the defined training steps. We also notice that the combination between $\epsilon$ \textit{and} $\delta$ is particularly helpful for spheres. The algorithms trained with $\beta$ especially struggle to grasp spheres. Furthermore, the reward framework $\epsilon$ initially trains faster than the reward frameworks that include the force agnostic metric $\delta$. Lastly, we recognize that the \textit{Hold Success} and \textit{Lift Success} graphs in Fig. \ref{fig:train_reward_frameworks} are very similar.

\begin{figure}[t]
	\centering
    	\includegraphics[width=\linewidth]{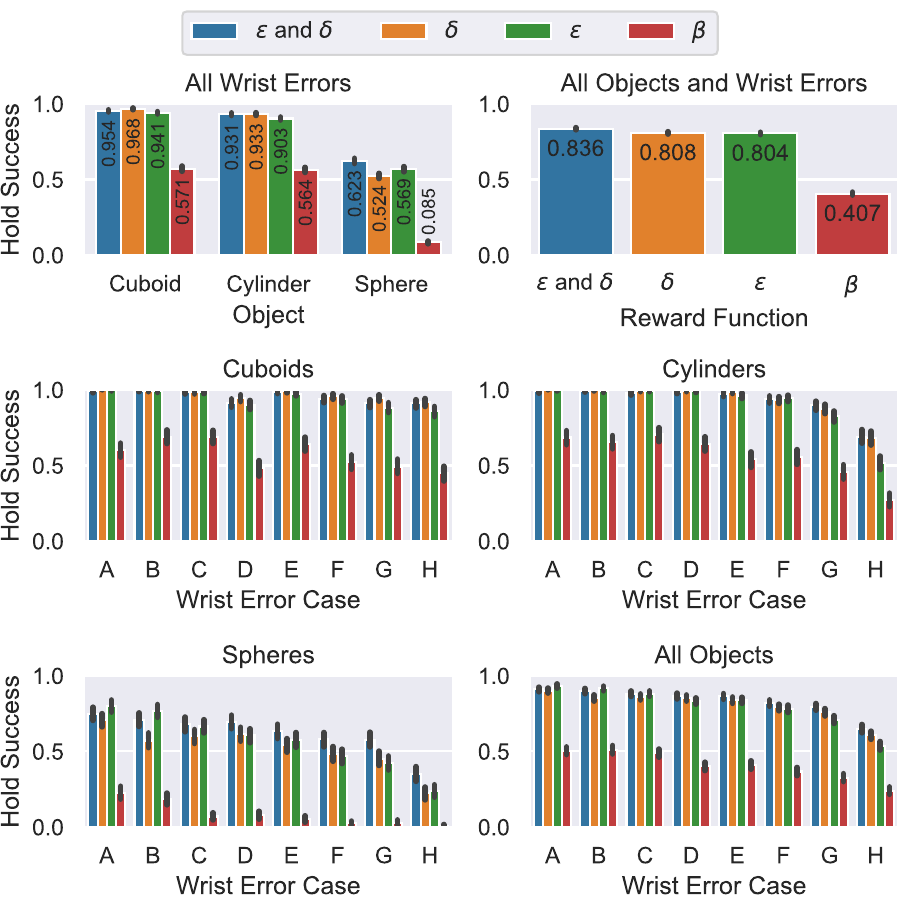}
    	\caption{Test results for reward frameworks.}
    	\label{fig:test_reward_framework}
\end{figure}

Fig. \ref{fig:test_reward_framework} summarizes the test results. All test results in this paper stem from 38400 grasps (40 models with different seeds $\times$ 4 frameworks $\times$ 240 test cases). Our main observation is that combining the geometric grasp stability metric $\epsilon$ with the force-agnostic metric $\delta$ yields the highest average success rates of 83.6\% across all objects (95.4\% for cuboids, 93.1\% for cylinders, and 62.3\% for spheres) over all wrist errors. The  $\epsilon$ \textit{and} $\delta$ framework outperforms the binary reward framework $\beta$ by 42.9\%. As expected, performance decreases for larger wrist errors. We show results of a one-sided, paired t-test in Table \ref{tab:t_test_reward} (mean of framework $x$ is $\mu_x$ and `$\approx$ 0.0' means that value was numerically zero).

\begin{table}[h]
\centering
\caption{Results of t-test for reward comparison.}
\begin{tabular}{|l|l|l|l|}
\hline
\textbf{Result} & $\mu_{\epsilon \text{ \textit{and} } \delta} >\mu_{\delta} $ & \begin{tabular}[c]{@{}l@{}}$\mu_{\epsilon \text{ \textit{and} } \delta} > \mu_{\epsilon} $\end{tabular} & \begin{tabular}[c]{@{}l@{}}$\mu_{\epsilon \text{ \textit{and} } \delta} > \mu_{\beta}  $\end{tabular} \\ \hline
p-value        & 3.1681 $10^{-10}$                                            & 2.0510 $10^{-12}$                                                                                  & $\approx$ 0.0                                                                                              \\ 
\hline
\end{tabular}
\label{tab:t_test_reward}
\end{table}

\subsection{Discussion}

This study investigates the tactile sensing needs in the reward of RL grasping controllers by incorporating highly accurate contact information via analytic grasp stability metrics. From the results of the t-test in Table \ref{tab:t_test_reward}, we conclude that the claim `the combination of $\epsilon$ \textit{and} $\delta$ outperforms all other tested rewards frameworks' is statistically significant ($p<0.01$ for all comparisons). The results demonstrate that information about contact positions and normals encoded in $\epsilon$ combines well with the force-based information in the $\delta$ reward. This result motivates building physical robotic hands capable of sensing these types of information. The low success rates for the spheres may be because they can roll and are therefore harder to grasp (cuboids and cylinders move comparatively less when touched by fingers or the palm). The observation that success rates after the \textit{lift} and the \textit{hold} stage are almost identical means that once the hand successfully lifts the object, the grasp is usually also stable enough to keep the object in hand until the very end of the grasp refinement episode.

The $\beta$ framework performs worst after the defined number of training steps, which is unsurprising because shaped rewards are known to be more sample efficient than sparse rewards  \cite{ng1999policyinvariance}. The $\beta$ framework may not constitute the best-performing alternative that is not based on analytic techniques from grasp analysis. However, it should be considered as a non-tactile reward baseline often used in related works \cite{merzic2019contact, wu2019mat}. Furthermore, the performance of the $\beta$ framework in Fig. \ref{fig:train_reward_frameworks} continues to rise slowly, and it would be interesting to evaluate at which success rates it plateaus.

\begin{figure}[t]
	\centering
    	\includegraphics[width=\linewidth]{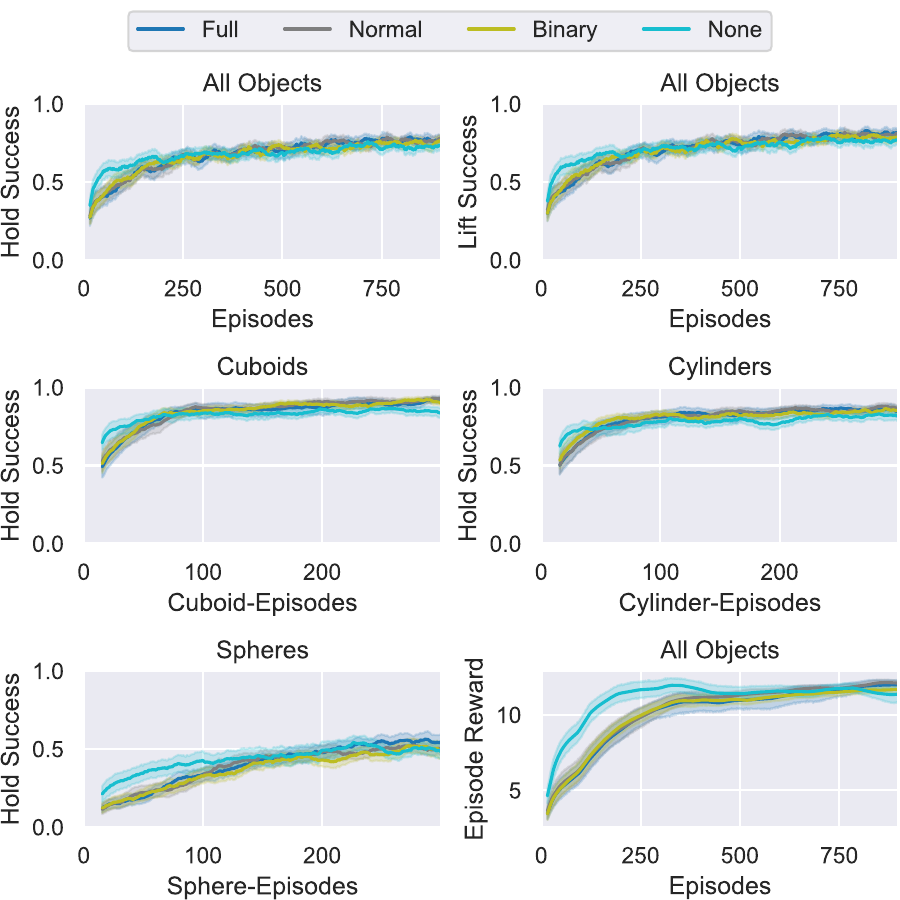}
    	\caption{Training results for contact sensing frameworks.}
    	\label{fig:train_force_frameworks}
\end{figure}

\section{Tactile Sensing and the State Vector\label{sec:contact_sensing}}

\subsection{Experimental Setup\label{sec:force_frameworks}}

In a second experiment, we investigate the effect of contact sensing resolution in the state vector on grasp refinement. We compare four contact sensing frameworks. The \textit{full} contact sensing framework receives the same state vector $\boldsymbol{s} \in \mathbb{R}^{70}$ as in section \ref{sec:state_action_space}. In the \textit{normal} framework, we only provide the algorithm with the contact normal forces and omit the tangential forces ($\boldsymbol{s} \in \mathbb{R}^{56}$). In the \textit{binary} framework we only give a binary signal whether a link is in contact (1) or not (0) ($\boldsymbol{s} \in \mathbb{R}^{56}$). Finally, we solely provide the joint positions in the \textit{none} framework ($\boldsymbol{s} \in \mathbb{R}^{7}$). We adjust the size of the input layer of the neural network from section \ref{sec:state_action_space} to match the size of the state vector of each framework. We keep the rest of the network's architecture fixed to allow a fair comparison. The reward function in these experiments is $\epsilon$ \textit{and} $\delta$ from Fig. \ref{fig:overview}. Hence, all contact sensing frameworks receive contact information indirectly via the reward. 

\subsection{Results}

Fig. \ref{fig:train_force_frameworks} shows the training performance of the contact sensing frameworks. Note that the \textit{full} framework is the same as the $\epsilon$ \textit{and} $\delta$ framework from section \ref{sec:reward_design}. We can observe that the \textit{none} framework initially learns faster than the other frameworks. However, after approximately 250 episodes, the frameworks that receive contact feedback outperform the \textit{none} framework, which plateaus at a lower success rate. 

\begin{figure}[t]
	\centering
    	\includegraphics[width=\linewidth]{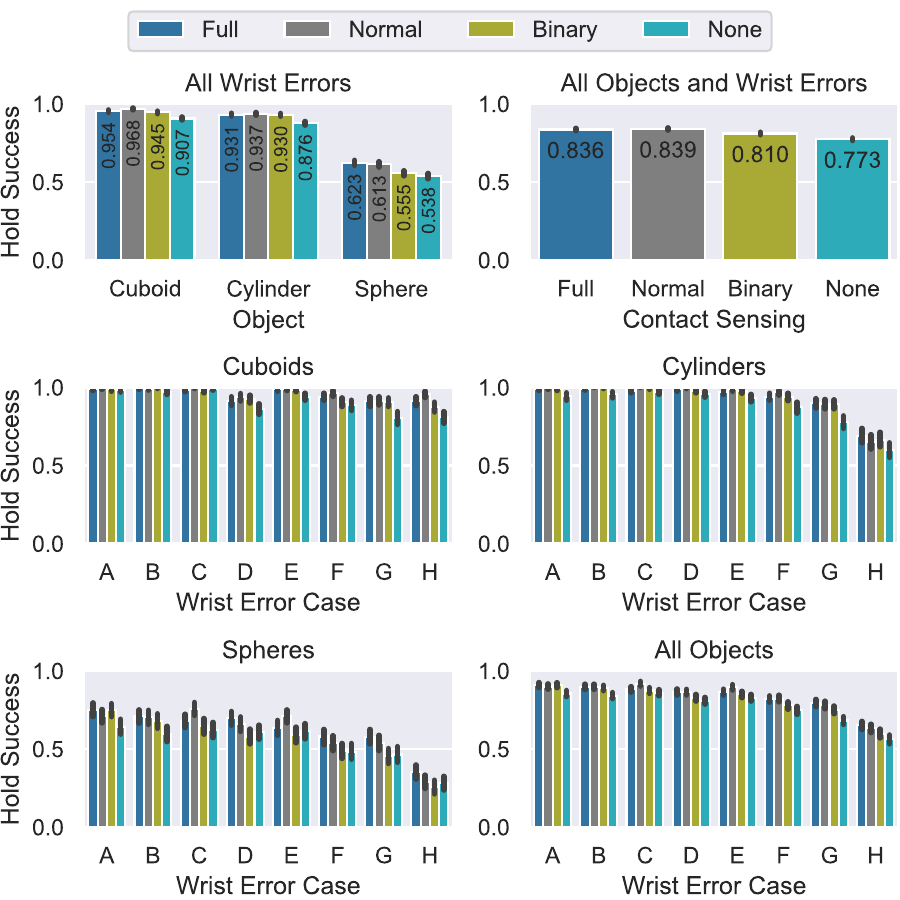}
    	\caption{Test results for contact sensing frameworks.}
    	\label{fig:test_force_framework}
\end{figure}

Fig. \ref{fig:test_force_framework} compares the test results of the different contact sensing frameworks. We observe that the frameworks which receive contact feedback (\textit{full}, \textit{normal}, \textit{binary}) outperform the \textit{none} framework by 6.3\%, 6.6\% and 3.7\%, respectively. Providing the algorithm with \textit{normal} force information yields a performance increase of 2.9\% compared to the \textit{binary} contact sensing framework. However, training with the \textit{full} contact force vector only increases the performance by 2.6\% compared to the \textit{binary} framework. Furthermore, the success rates for cuboids and cylinders are higher than for spheres (for the \textit{normal} force framework the success rates are 96.8\%, 93.7\%, 61.3\%, respectively). We show the results of a one-sided, paired t-test in Table \ref{tab:t_test_force_frameworks}. 

\begin{table}[h]
\centering
\caption{Results of t-test for contact sensing comparison.}
\begin{tabular}{|l|l|l|l|}
\hline
\textbf{Result} & $\mu_{\text{\textit{normal}}} > \mu_{\text{\textit{full}}}$ & \begin{tabular}[c]{@{}l@{}}$\mu_{ \text{\textit{normal}}}  >  \mu_{\text{\textit{binary}}} $\end{tabular} & $\mu_{\text{\textit{normal}}} > \mu_{\text{\textit{none}}}$ \\ \hline
p-value        & 0.2232                           & 7.0177 $10^{-11}$                                                                       & 1.3087 $10^{-46}$                 \\ \hline
\end{tabular}
\label{tab:t_test_force_frameworks}
\end{table}

\subsection{Discussion}
This experiment studies how contact sensing resolution in the policy's state vector is related to grasp success when training with fully contact informed rewards. Thereby, we investigate the viability of our hypothesized training and deployment workflow in Figure \ref{fig:c2_flow}. The training curves of the \textit{full}, \textit{normal} and \textit{binary} frameworks in Fig. \ref{fig:train_force_frameworks} are hard to distinguish, which indicates a similar training performance in all cases. Each data point in the training curves includes the outcome of only one grasp refinement episode per model (one object $O$ and one wrist error $W$). This punctual evaluation poorly reflects on the \textit{overall} model performance. Therefore, we should focus our analysis on the test results from the 240 experiments per model over multiple objects and wrist errors which provide a more comprehensive model evaluation. In the test results, we observe statistically significant improvements for the \textit{normal} force framework when compared to the \textit{binary} and \textit{none} frameworks (p-values in Table \ref{tab:t_test_force_frameworks} $< 0.01$). However, these improvements are small, and the results suggest that an affordable \textit{binary} contact sensor suite may be suitable if a small decrease in performance is tolerable. The surprisingly good performance of the \textit{none} framework means that agents can refine grasps solely based on the crude contact feedback of finger joint position data when trained with rewards that encode grasp stability. These results support our hypothesis that RL grasping algorithms are deployable to hands with reduced contact sensor resolution at little performance decrease when incorporating rich tactile feedback at train time. These results also have implications for systems trained in simulation that usually suffer from a sim-to-real gap. The gap may be reduced by choosing only a feature set in the state vector that is cheaply and accurately available on the real hand (e.g., joint positions) and integrating harder-to-obtain information in the reward, which is easily computable in simulation. 

Interestingly, the algorithms trained with the \textit{full} force vector perform approximately on par with the ones that receive the \textit{normal} force information (the small difference in success rates of 0.3\% is not statistically significant because p-value $> 0.01$ in Table \ref{tab:t_test_force_frameworks}). This observation is counterintuitive since tangential forces are prominent in the designed grasping task. This result could be due to three reasons. 
(1) The \textit{full} force framework is the framework with the largest state vector (see section \ref{sec:force_frameworks}) and therefore requires the most training data because it has the most network parameters. Future experiments should run more training steps. 
(2) The models trained with the \textit{full} framework will have to internally represent the concept of the friction cone, which may be a complex notion to learn from discontinuous contact data (sometimes there is contact on a link, sometimes there is not). An alternative representation of the tangential forces could be an exciting avenue for research (e.g., provide margin to the friction cone instead of tangential force vector). 
(3) Lastly, contact forces in simulated environments are known to be unstable \cite{hsu2014DARPA}, especially when simulating robotic grasping \cite{taylor2016grasp}. Hence, another reason for our observation may be that since simulated contact forces are not always physically meaningful, they may not necessarily constitute a good proxy of grasp success in simulation. 

We relate the differences in learning speed to the size of the state vector. The \textit{none} framework has a smaller state vector and can hence learn faster, while the frameworks that process contact information require more training data to converge. The relative performance decrease when reducing contact sensor resolution is approximately the same across all objects, even though they have different geometries and grasping strategies. This result suggests that our conclusions are representative of a variety of object geometries.

\section{Conclusion \label{sec:conclusion}}

This paper investigated the importance of tactile signals in the reward and the policy's state vector to identify the tactile sensing needs in RL-based grasping algorithms. We found that rewards incorporating contact positions, normals, and forces are the most powerful optimization objectives for RL grasp refinement controllers. While this tactile information is essential in the reward function, we uncovered that reducing contact sensor resolution in the policy's state vector decreases algorithm performance only by a small amount. This result has implications for the design of physical grippers and their training and deployment workflows.

There are several exciting directions for future work. Following our first conclusion, it is essential to build a physical robotic hand with advanced sensing capabilities to calculate grasp metrics. Secondly, we aim to test the proposed training and deployment workflow, providing only limited contact information in the state vector and testing the algorithm on other robotic hands. Our second experiment mainly examined the effect of the representation of contact forces on grasp refinement. Therefore, future ablation studies should quantify the relevance of contact normal and position sensing in the state vector.

\bibliographystyle{IEEEtran}
\bibliography{bibliography}
\end{document}